\documentclass[sigconf]{acmart}

\usepackage{multicol}
\usepackage{multirow}
\usepackage[table]{xcolor}
\definecolor{TableAccent}{HTML}{E5F0FA}
\AtBeginDocument{%
  }

\setcopyright{acmlicensed}
\copyrightyear{2026}
\acmYear{2026}
\acmConference[MM '26]{Proceedings of the 34th ACM International Conference on Multimedia}{November 10--14, 2026}{Rio de Janeiro, Brazil}
\acmBooktitle{Proceedings of the 34th ACM International Conference on Multimedia (MM '26), November 10--14, 2026, Rio de Janeiro, Brazil}
\begin{document}

\title{Spatial-Temporal Decoupled Reference Conditioning for Identity-Preserving Text-to-Video Generation}

\author{Yuheng Chen}
\authornote{Both authors contributed equally to this research.}
\affiliation{%
  \institution{Shanghai Jiao Tong University}
  \city{Shanghai}
  \state{Shanghai}
  \country{China}
}
\email{cyh627@sjtu.edu.cn}

\author{Teng Hu}
\authornotemark[1]

\affiliation{%
  \institution{Shanghai Jiao Tong University}
  \city{Shanghai}
  \state{Shanghai}
  \country{China}
}
\email{hu-teng@sjtu.edu.cn}

\author{Yuji Wang}
\affiliation{%
  \institution{Shanghai Jiao Tong University}
  \city{Shanghai}
  \state{Shanghai}
  \country{China}
}
\email{yujiwang@sjtu.edu.cn}

\author{Qingdong He}
\affiliation{%
  \institution{University of Electronic Science and Technology of China}
  \city{Chengdu}
  \state{Sichuan}
  \country{China}
}
\email{heqingdong@alu.uestc.edu.cn}

\author{Lizhuang Ma}
\authornote{Corresponding authors.}
% \authornotemark[2]
\affiliation{%
  \institution{Shanghai Jiao Tong University}
  \city{Shanghai}
  \state{Shanghai}
  \country{China}
}
\email{ma-lz@cs.sjtu.edu.cn}

\author{Jiangning Zhang}
\authornote{Project lead.}
% \authornotemark[3]
\affiliation{%
  \institution{Zhejiang University}
  \city{Hangzhou}
  \state{Zhejiang}
  \country{China}
}
\email{186368@zju.edu.cn}

% \affiliation{%

% \affiliation{%

% \affiliation{%

\renewcommand{\shortauthors}{Chen et al.}

% \begin{abstract}
%   A clear and well-documented \LaTeX\ document is presented as an
%   article formatted for publication by ACM in a conference proceedings
%   or journal publication. Based on the ``acmart'' document class, this
%   article presents and explains many of the common variations, as well
%   as many of the formatting elements an author may use in the
%   preparation of the documentation of their work.
% \end{abstract}
\begin{abstract}
Identity-preserving video generation (IPVG) aims to synthesize high-fidelity videos that follow text prompts while faithfully preserving a reference identity. 
Despite recent progress, existing IPVG methods still struggle to balance high-level semantic control and low-level identity fidelity. 
To bridge this gap, we propose \textbf{ST-DRC}, an effective \textbf{S}patial-\textbf{T}emporal \textbf{D}ecoupled \textbf{R}eference \textbf{C}onditioning framework for identity-preserving text-to-video generation. 
At the framework level, ST-DRC performs latent in-context feature injection by encoding the reference image with the video VAE and concatenating it with noisy video latents, enabling rich low-level identity details to be accessed without additional adapters. 
To separate identity-aware reference retrieval from appearance copying, we introduce \textbf{TASS-RoPE}, a Temporal-Adjacent Spatial-Shifted RoPE scheme that places reference tokens near the video sequence in time but shifts them in space, allowing reference information to flow through spatio-temporal attention while suppressing pixel-level copy-paste shortcuts. 
To further prevent shortcut learning and strengthen the otherwise diluted identity supervision in the diffusion objective, we combine appearance-invariant reference augmentation with face-guided identity objectives, encouraging the model to preserve identity under variations in color, pose, and layout. 
At inference time, we introduce a three-stream reference classifier-free guidance strategy that independently controls text adherence and reference fidelity. 
Experiments demonstrate that ST-DRC achieves strong identity preservation, prompt alignment, temporal consistency, and video quality with a lightweight design built on LTX-2.3. 
Our method ranks among the top submissions in the facial identity-preserving video generation track, validating the effectiveness of spatial-temporal decoupled reference conditioning. 
Code is available at \url{https://github.com/AliothChen/ST-DRC}.
\end{abstract}



\begin{teaserfigure}
  \includegraphics[width=\textwidth]{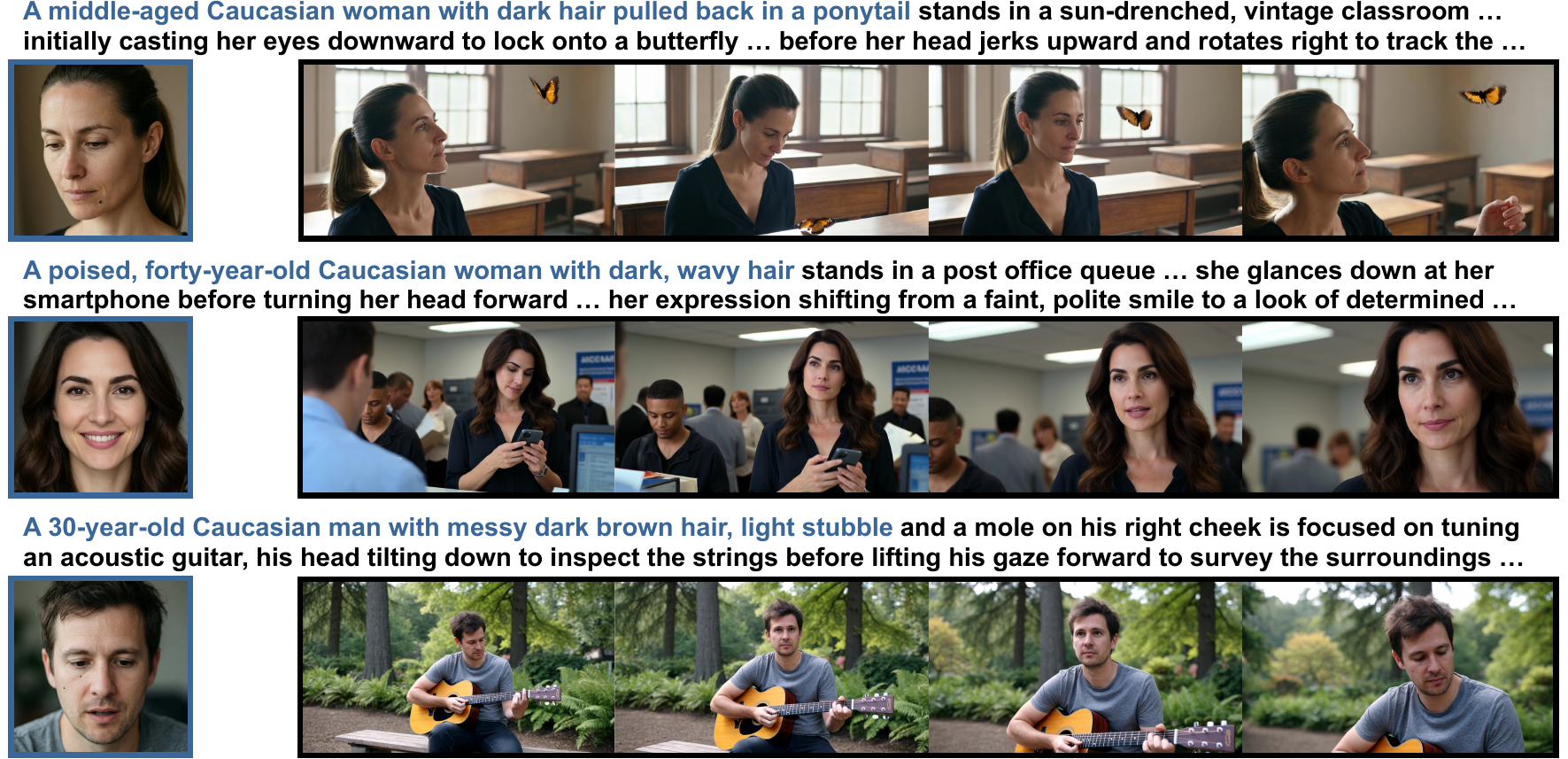}
\caption{
Examples of identity-preserving text-to-video generation by our ST-DRC. 
Given a reference face, our method generates high-fidelity videos with consistent identity. 
Blue text and boxes indicate the reference identity.
}
  % \Description{Enjoying the baseball game from the third-base
  % seats. Ichiro Suzuki preparing to bat.}
  \label{fig:teaser}
\end{teaserfigure}


\maketitle

\section{Introduction}

Text-to-video generation has advanced rapidly with large-scale diffusion transformers~\cite{dit}, enabling visually compelling videos from open-ended text prompts~\cite{wan,hunyuanvideo,hunyuanvideo-1.5,skyreels-v2,cogvideox,mova,ltx-2,ovi,harmony,polyvivid,ultragen,hu2025hunyuancustom,t3video}. 
However, text alone is insufficient when the generated video is expected to depict a specific identity. 
Identity-preserving video generation (IPVG)~\cite{jiface,consistid,ipvg2501,ipvg2502} addresses this limitation by conditioning video synthesis on both a reference image and a text prompt, requiring the generated video to follow the prompt while maintaining the reference identity across time. 
% This requirement is particularly demanding in the facial setting: identity depends on subtle local structures, proportions, and textures, while the video should still allow natural changes in pose, expression, illumination, motion, and scene context. 
% Thus, the central challenge is not simply to inject a reference image, but to preserve identity-relevant cues without copying irrelevant appearance factors.

IPVG involves \textbf{\textit{two}} tightly coupled challenges: 
\textbf{\textit{high-level prompt controllability}}, where the generated video should follow the text-specified action, scene, style, and temporal dynamics, and 
\textbf{\textit{low-level identity fidelity}}, where identity-bearing facial details from the reference image should be preserved without carrying over nuisance appearance factors such as lighting, head pose, background, or camera layout. 
Existing IPVG methods can be roughly grouped into \textbf{\textit{two}} paradigms. 
\textbf{\textit{1) Semantic reference injection.}} 
Adapter- or embedding-based methods~\cite{consistid,uniportrait,lynx,ip-adapter,jiface,magicmirror,id-animator} inject the reference through compact semantic features, face embeddings, or additional cross-attention branches. 
This paradigm is compatible with text-conditioned generation and provides robust global guidance, but the compressed reference representation may discard fine-grained spatial details that are crucial for facial identity preservation. 
\textbf{\textit{2) Latent reference injection.}} 
Another line of methods~\cite{phantom,kaleido} encodes the reference image into the same latent space as video frames and concatenates it with video tokens, allowing the model to access richer spatial details through spatio-temporal attention. 
While more promising for identity fidelity, this paradigm also exposes the model to nuisance appearance factors in the reference image, such as pose, illumination, background, and layout. 
This leads to an appearance copy-paste~\cite{echovideo,phantom-data} problem, where details irrelevant to identity are mistakenly transferred to the generated video, causing appearance leakage, ghosting artifacts, reduced motion diversity, and weaker prompt following.

To address this issue, we propose \textbf{ST-DRC}, a spatial-temporal decoupled reference conditioning framework built on LTX-2.3~\cite{ltx-2}. 
Our method first encodes the reference image with the video VAE and concatenates it with noisy video latents as a non-decoded identity memory, allowing the model to access rich spatial identity cues through its native spatio-temporal attention. 
To prevent this memory from being treated as a physical frame or a pixel-aligned template, we introduce \textbf{TASS-RoPE}, which places reference tokens temporally adjacent to the video sequence while shifting them away in the spatial dimensions. 
This design keeps reference information close enough for attention-based identity retrieval, but breaks direct spatial alignment that would otherwise encourage low-level copying. 
During training, we further perturb the reference image in geometry, color, and layout, making non-identity appearance factors unreliable. 
Since such decoupling mainly removes shortcuts, we complement it with a face-guided identity objective that directly supervises the generated faces in an identity-discriminative embedding space, alleviating the dilution of sparse facial identity signals under the global diffusion objective. 
At inference time, a three-stream reference classifier-free guidance strategy~\cite{cfg} independently adjusts prompt adherence and reference fidelity.

Our contributions are summarized as follows:\\
\hspace*{1em}\textbf{\textit{1)}} We propose \textbf{ST-DRC}, a unified latent reference conditioning framework that encodes the reference image with the video VAE and concatenates it with noisy video latents as a non-decoded identity memory, enabling rich low-level identity details to be accessed without additional adapters.\\
\hspace*{1em}\textbf{\textit{2)}} We introduce TASS-RoPE, a temporal-adjacent and spatial-shifted positional design that enables effective reference attention while suppressing pixel-level spatial copying.\\
\hspace*{1em}\textbf{\textit{3)}} We combine appearance-invariant reference perturbation with face-guided identity supervision, jointly removing nuisance appearance shortcuts and strengthening identity learning.\\
\hspace*{1em}\textbf{\textit{4)}} We introduce three-stream reference guidance for inference-time control over prompt adherence and reference fidelity, achieving strong identity preservation and temporal consistency.

\begin{figure*}
    \centering
    \includegraphics[width=\linewidth]{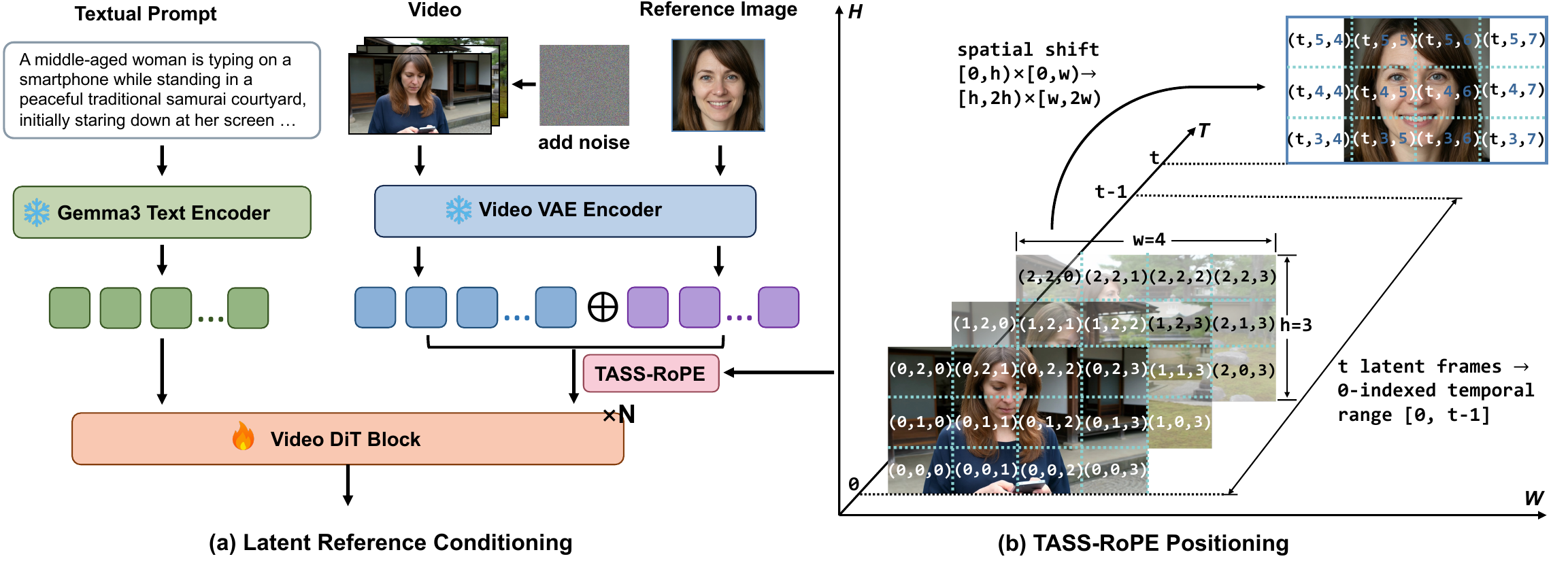}
        % \vspace{-7mm}
\caption{
Overview of ST-DRC. 
(a) The reference image is encoded into the video latent space and concatenated with noisy video latents as a non-decoded identity memory. 
(b) TASS-RoPE places reference tokens near the video sequence in time but shifts them in space, enabling identity retrieval while reducing pixel-level copy-paste.
}
    \label{fig:method}
    % \vspace{-2mm}
\end{figure*}

\section{Related Works}

% \subsection{Identity-preserving video generation}
\noindent\textbf{Identity-preserving video generation.}
Identity-preserving video generation (IPVG) aims to synthesize videos that follow textual instructions while maintaining the identity of a reference subject. 
% Compared with general text-to-video generation, IPVG imposes a stricter constraint for identity consistency. 
Existing methods typically rely on specialized identity-conditioning mechanisms. 
Adapter- or embedding-based methods inject compact reference features through face encoders, identity embeddings, cross-attention branches, or multimodal fusion modules~\cite{ip-adapter,id-animator,consistid,lynx,magicmirror,echovideo,humo}. 
They improve identity control and integrate well with text-conditioned generation, but often introduce additional identity-specific modules and may discard fine-grained spatial details through feature compression. 
Another line of methods encodes reference images into the video latent space and injects them as visual tokens, preserving richer spatial details for spatio-temporal attention~\cite{videobooth,phantom,kaleido,vace}. 
However, the reference latent also carries pose, illumination, background, and layout, which can cause an appearance copy-paste problem where identity-irrelevant details are mistakenly transferred to the generated video. 
Multi-stage, training-free, and post-hoc optimization methods further improve identity preservation through staged customization, prompt/reference enhancement, guidance design, reward optimization, or multiple fine-tuned experts~\cite{dreamvideo,customvideo,videodreamer,still-moving,customcrafter,ipvg2501,identitygrpo,ipvg2502}. 
These methods are effective but usually require per-subject adaptation, extra inference-time processing, or additional optimization stages. 
In contrast, our method represents the reference face directly in the video latent space as a non-decoded identity condition, avoiding additional identity encoders and per-identity tuning while addressing appearance entanglement through spatial-temporal decoupling, reference augmentation, and reference-conditioned guidance.
% In contrast, our method uses the reference face as a non-decoded latent condition, avoiding extra identity encoders or per-identity tuning while mitigating appearance entanglement.

% \subsection{Subject-to-video generation}
\noindent\textbf{Subject-to-video generation.}
Subject-to-video (S2V) generation conditions a video generation model on one or more reference images of target subjects, covering broader categories such as persons, objects, animals, and backgrounds. 
Early personalization methods such as DreamBooth~\cite{dreambooth} and DreamVideo~\cite{dreamvideo} achieve subject fidelity through subject-specific or few-shot customization, but their per-subject optimization limits scalability. 
Recent open-set or zero-shot methods improve subject consistency through adapter-based conditioning~\cite{videoalchemist}, cross-attention injection~\cite{conceptmaster,polyvivid,customcrafter}, anchored prompts~\cite{movieweaver}, LoRA-based customization~\cite{firstframe,videomage}, latent reference concatenation~\cite{contextanyone,hu2025hunyuancustom,Omni-Customizer}, or multimodal conditioning~\cite{bindweave,smrabooth,skyreels-a2,animateanyone2,videojam}. 
Large-scale datasets and benchmarks such as OpenS2V~\cite{yuanopens2v}, Phantom-Data~\cite{phantom-data}, and OpenSubject~\cite{opensubject} further support evaluation and training for subject-consistent generation. 
Notably, recent S2V studies highlight persistent challenges such as multi-subject inconsistency, background entanglement, reduced reference fidelity, semantic drift, and copy-paste artifacts under reference-image conditioning~\cite{kaleido,phantom-data}. 
Although these works provide important designs for reference-conditioned video generation, their objective is broader than facial IPVG, where identity preservation depends on fine-grained facial cues while pose, expression, lighting, and scene context should remain flexible. 
% Our method follows the latent-concatenation direction, but specializes it for facial IPVG by treating the reference portrait as a non-temporal identity memory with TASS-RoPE, appearance-invariant augmentation, and reference-conditioned guidance.
Our method follows latent concatenation, but tailors it to facial IPVG with TASS-RoPE, appearance-invariant augmentation, and reference-conditioned guidance.

\section{Methods}
ST-DRC generates identity-preserving videos conditioned on a reference image $I_{\mathrm{ref}}$ and a text prompt $y$. 
Given a noisy video latent $z_t$, our method injects the reference identity into the denoising process while preserving prompt-driven motion and scene dynamics.  
We implement ST-DRC on the \textbf{\textit{video branch}} of LTX-2.3~\cite{ltx-2}, and exclude its audio branch since our setting does not involve audio input or output. 
As shown in Fig.~\ref{fig:method}, we first encode $I_{\mathrm{ref}}$ into the video VAE latent space and append it to the video latent sequence as a non-decoded identity memory. 
The reference tokens are then assigned Temporal-Adjacent Spatial-Shifted RoPE (TASS-RoPE) coordinates, enabling identity information to be retrieved through spatio-temporal attention while discouraging spatial copy-paste. 
During training, reference augmentation and face-guided identity supervision further suppress nuisance appearance shortcuts and strengthen identity learning. 
At inference time, three-stream reference CFG provides independent control over prompt adherence and reference fidelity.

\subsection{Latent In-Context Reference Injection}
\label{sec31}

Our first design is to represent the reference image as an in-context latent memory rather than an additional semantic embedding or a decoded video frame. 
Since the video branch of LTX-2.3 operates in a VAE latent space with spatio-temporal attention, the reference image can be encoded into the same latent space as the video frames and jointly processed with noisy video tokens.
Before encoding, we resize the reference image to the target video resolution by scaling it with the largest factor that preserves its aspect ratio, followed by symmetric padding on the shorter side to match the spatial size of the video frames.
Let $E_{\mathrm{vae}}$ denote the video VAE encoder. 
We encode the reference image as:
\begin{equation}
    z_{\mathrm{ref}} = E_{\mathrm{vae}}(I_{\mathrm{ref}}),
\end{equation}
where $z_{\mathrm{ref}} \in \mathbb{R}^{1 \times H \times W \times C}$ denotes the reference latent. 
As shown in Fig.~\ref{fig:method}(a), given the noisy video latent $z_t \in \mathbb{R}^{T \times H \times W \times C}$, we append the reference latent along the temporal dimension:
\begin{equation}
    \tilde z_t = [z_t, z_{\mathrm{ref}}],
\end{equation}
where $\tilde z_t \in \mathbb{R}^{(T+1) \times H \times W \times C}$.
After concatenation, both video and reference latents are forwarded through the Video DiT block, allowing the reference latent to provide low-level visual details via spatio-temporal attention. 
The output corresponding to $z_{\mathrm{ref}}$ is discarded, meaning that it is neither decoded into video frames nor included in the training loss.

\subsection{Temporal-Adjacent Spatial-Shifted RoPE}
\label{sec32}

Latent concatenation alone is insufficient for effective reference conditioning, because the Video DiT models video latents with 3D RoPE~\cite{su2024roformer} in self-attention~\cite{transformer}. 
If no dedicated RoPE coordinates are assigned to the reference latent, its tokens collapse to identical or default positions and may overlap with the first video frame, leading to suboptimal position modeling and undesired interference with temporal attention. 
Therefore, after appending the reference latent, we further specify its temporal and spatial coordinates in the 3D RoPE space. 
% The goal is to keep the reference close enough for identity retrieval, while preventing it from being interpreted as a real video frame or a spatially aligned template.

Assume the video latent contains $T$ frames with temporal indices $0,\ldots,T-1$ and spatial indices $(i,j)$, where $0 \leq i < H$ and $0 \leq j < W$. 
For video tokens, we keep the standard 3D coordinates:
\begin{equation}
    p_v(t,i,j) = (t,i,j), \quad t=0,\ldots,T-1.
\end{equation}
For reference tokens, we propose \textbf{Temporal-Adjacent Spatial-Shifted RoPE} (TASS-RoPE) as shown in Fig.~\ref{fig:method}(b), which assigns:
\begin{equation}
    p_r(i,j) = \mathrm{TASS}(i,j) = (T, H+i, W+j).
\end{equation}
Thus, the reference latent occupies the first out-of-frame temporal slot and a non-overlapping spatial region $[H,2H)\times[W,2W)$.

The temporal coordinate is chosen as $T$ for \textbf{\textit{two}} reasons. 
\textbf{\textit{First}}, it lies outside the video frame range $[0,T-1]$, so the reference tokens do not collide with any real video frame or disturb the original temporal ordering among video tokens. 
\textbf{\textit{Second}}, it is length-adaptive. 
A fixed absolute temporal index, such as $100$, is undesirable because the same reference position induces inconsistent relative-position patterns across videos of different lengths, making training less stable. 
We also avoid placing the reference at a remote index such as $T+100$, since RoPE induces distance-decaying inter-token dependency, and an excessively large temporal gap may weaken the transfer of fine-grained visual details from the reference tokens to the video tokens. 
Instead, we use $T$, the nearest out-of-frame temporal position, which avoids collision with video frames while keeping the reference temporally close to the entire video sequence.

For the spatial coordinates, we shift the reference latent from the original video region $[0,H)\times[0,W)$ to $[H,2H)\times[W,2W)$. 
This keeps the internal spatial geometry of the reference image unchanged, so neighboring facial regions in the reference latent still preserve their relative spatial relationships. 
Meanwhile, no reference token shares the same spatial coordinate with any video token. 
This spatial non-overlap discourages the model from learning a coordinate-aligned \textit{copy-paste shortcut}, where local appearance details such as pose, lighting, background, or layout are directly transferred from the reference image to the generated video.

% Overall, TASS-RoPE makes the reference latent temporally adjacent but spatially shifted. 
% The reference remains close in time to support identity retrieval through spatio-temporal attention, while the spatial shift prevents direct pixel-level alignment. 
% This positional design turns latent reference conditioning from frame-like copying into memory-like reference retrieval, retaining low-level identity details while suppressing appearance leakage.

\begin{figure*}[t]
    \centering
    \includegraphics[width=\linewidth]{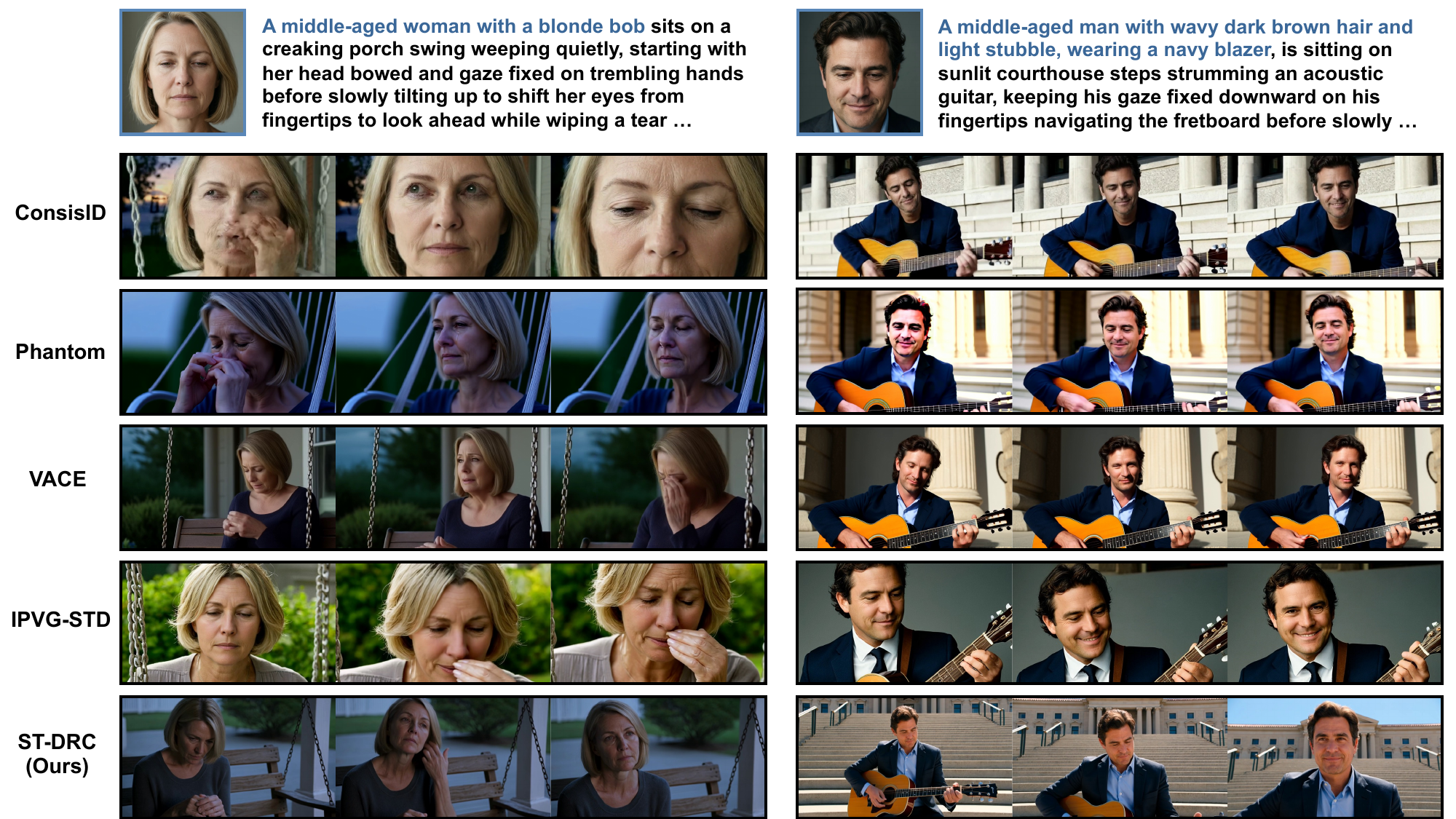}
        % \vspace{-7mm}
\caption{
Qualitative comparison on VIP-200K~\cite{ipvg}. 
Given the same reference image and text prompt, ST-DRC preserves the target identity more faithfully while producing prompt-consistent and visually coherent videos compared with baselines.
}
\label{fig:compare}
% \vspace{-3mm}
\end{figure*}

% \subsection{Reference-Robust Identity Training}
% \label{sec33}
\subsection{Reference-Robust Identity Enhancement}
\label{sec33}

To further improve \textbf{\textit{model capability}} and \textbf{\textit{training efficiency}}, we respectively introduce the following \textbf{\textit{two}} complementary training strategies:

\textbf{\textit{1) Image-level reference augmentation.}}
During training, we randomly apply identity-preserving augmentations to the reference image before VAE encoding:
\begin{equation}
    z_{\mathrm{ref}} = E_{\mathrm{vae}}(\mathcal{A}(I_{\mathrm{ref}})),
\end{equation}
where $\mathcal{A}$ is a stochastic augmentation operator sampled with predefined probabilities from \textbf{\textit{i)}} geometric transformations, such as horizontal flipping, slight rotation, and mild spatial cropping, and \textbf{\textit{ii)}} photometric transformations, such as color jittering. 
At inference time, no augmentation is applied and the original reference image is used. 
This strategy further mitigates the copy-paste problem and encourages the model to learn more robust high-level identity features instead of relying on low-level appearance shortcuts.

\textbf{\textit{2) Face-guided auxiliary identity loss.}}
% As illustrated in Fig.~\ref{fig:ltx_wan}, applying the latent reference injection (Sec.~\ref{sec31}) and TASS-RoPE (Sec.~\ref{sec32}) to different backbones leads to different behaviors. 
% \textit{LTX-2.3 produces valid videos, while Wan degenerates into severely degraded results. }
% This observation suggests that our conditioning form is \textbf{\textit{in-domain}} for LTX-2.3, so the standard flow-matching loss~\cite{flowmatching,rectifiedflow,sd3} provides only weak supervision for guiding the model to learn identity information from the reference. 
% Moreover, since the loss is averaged over all video tokens, sparse facial identity cues can be easily diluted by non-face tokens.
The standard flow-matching objective~\cite{flowmatching,rectifiedflow,sd3} supervises the model by regressing the velocity field over the entire video latent. 
While effective for overall video generation, this global objective provides limited direct supervision for learning reference identity. 
Since the loss is averaged over all video tokens, the effective identity-related loss is relatively small and can be easily diluted by numerous non-face tokens, resulting in less efficient identity training.

To provide a direct identity signal, we introduce a \textbf{\textit{face-guided auxiliary loss}} computed from a clean-latent estimate. 
Given the noisy video latent $z_t$ and the predicted velocity $\hat v_\theta$, we estimate the clean video latent as:
\begin{equation}
    \hat z_0 = z_t - t \hat v_\theta(\tilde z_t, t, y).
\end{equation}
We then decode the estimated clean latent using the frozen VAE decoder:
\begin{equation}
    \hat x^{(f)} = D_{\mathrm{vae}}(\hat z_0^{(f)}), 
    \quad f \in \mathcal{F},
\end{equation}
where $\mathcal{F}$ denotes the frame indices. 
InsightFace~\cite{insightface} is employed to detect face boxes, and the aligned face crops are passed to a frozen ArcFace encoder $\Phi$:
\begin{equation}
    e_f = \Phi(\operatorname{Align}(\hat x^{(f)})), 
    \quad
    e_{\mathrm{ref}} = \Phi(\operatorname{Align}(I_{\mathrm{ref}})).
\end{equation}
The reference embedding $e_{\mathrm{ref}}$ is computed from the original, unaugmented reference image and detached during training.

The identity loss is defined as the cosine distance in the ArcFace~\cite{arcface} embedding space:
\begin{equation}
    \mathcal{L}_{\mathrm{id}}
    =
    \frac{
    \sum_{f \in \mathcal{F}} m_f \left(1-\cos(e_f,e_{\mathrm{ref}})\right)
    }{
    \sum_{f \in \mathcal{F}} m_f + \epsilon
    },
\end{equation}
where $m_f$ indicates whether a valid face is detected in frame $f$. 
To reduce cross-frame identity drift, we further add a temporal identity consistency loss:
\begin{equation}
    \bar e =
    \frac{\sum_{f \in \mathcal{F}} m_f e_f}
    {\sum_{f \in \mathcal{F}} m_f + \epsilon},
    \quad
    \mathcal{L}_{\mathrm{tic}}
    =
    \frac{
    \sum_{f \in \mathcal{F}} m_f \left(1-\cos(e_f,\bar e)\right)
    }{
    \sum_{f \in \mathcal{F}} m_f + \epsilon
    }.
\end{equation}

Since the clean-latent estimate is more reliable at lower noise levels, we weight the auxiliary identity losses with an SNR-inspired coefficient:
\begin{equation}
    \mathcal{L}
    =
    \mathcal{L}_{\mathrm{flow}}
    +
    w_{\mathrm{aux}}(t)
    \left(
    \lambda_{\mathrm{id}} \mathcal{L}_{\mathrm{id}}
    +
    \lambda_{\mathrm{tic}} \mathcal{L}_{\mathrm{tic}}
    \right).
\end{equation}
For rectified-flow interpolation $z_t=(1-t)z_0+t\epsilon$, we define:
\begin{equation}
    \mathrm{SNR}(t)=\left(\frac{1-t}{t+\epsilon}\right)^2,
    \quad
    w_{\mathrm{aux}}(t)=
    \left(
    \frac{\mathrm{SNR}(t)}
    {\mathrm{SNR}(t)+1}
    \right)^{\gamma},
\end{equation}
where $\gamma$ controls the emphasis on low-noise timesteps. 
The coefficients $\lambda_{\mathrm{id}}$ and $\lambda_{\mathrm{tic}}$ balance identity preservation and temporal identity consistency.

\subsection{Decoupled Text-Reference Guidance}
\label{sec34}

To independently control prompt adherence and reference fidelity at inference time, we adopt a \textbf{\textit{decoupled text-reference}} classifier-free guidance strategy~\cite{cfg}. 
Specifically, we compute \textbf{\textit{three}} velocity predictions at each denoising step: \textbf{\textit{1)}} an unconditional prediction $\hat v_{\emptyset}$, \textbf{\textit{2)}} a text-only prediction $\hat v_{y}$, and \textbf{\textit{3)}} a text-reference prediction $\hat v_{y,r}$. 
The final guided velocity is defined as:
\begin{equation}
    \hat v_{\mathrm{cfg}}
    =
    \hat v_{\emptyset}
    +
    s_y \left(\hat v_y - \hat v_{\emptyset}\right)
    +
    s_r \left(\hat v_{y,r} - \hat v_y\right),
\end{equation}
where $s_y$ controls text guidance strength and $s_r$ controls reference guidance strength. 
Unless otherwise specified, we set $s_y=5.0$ and $s_r=7.5$ in all experiments. 
This formulation decomposes the guidance direction into a text direction $\hat v_y-\hat v_{\emptyset}$ and a reference direction $\hat v_{y,r}-\hat v_y$, allowing prompt following and identity preservation to be adjusted separately.

To enable these three prediction streams, we apply \textbf{\textit{independent condition dropout}} during training. 
The text condition is dropped with probability $p_y=0.05$, and the reference condition is dropped with probability $p_r=0.20$. 
% Since the two dropout events are sampled independently, the model observes four conditioning cases during training: unconditional, text-only, reference-only, and text-reference. 
% This makes the same model support decoupled guidance at inference without training separate conditional and unconditional networks.

\begin{table*}[t]
\centering
\caption{
Quantitative comparison on the VIP-200K test set~\cite{ipvg}. 
We compare ST-DRC with representative IPVG and S2V baselines, and report cumulative ablations from the LTX-2.3 base model to the full ST-DRC. 
Best results are marked in \textbf{bold}.
}
\setlength{\tabcolsep}{3.1pt}
\begin{tabular}{lcccccccccccc}
\toprule
\multirow{2}{*}{Method}    
& \multicolumn{5}{c}{Components}
& \multicolumn{2}{c}{Identity Preservation}  
& Prompt Alignment
& \multicolumn{4}{c}{Video Quality}          \\ 
\cmidrule(lr){2-6} \cmidrule(lr){7-8} \cmidrule(lr){10-13}
& Ref. 
& TASS 
& Aug. 
& ID 
& CFG
& FaceSim-Arc$\uparrow$ 
& FaceSim-Cur$\uparrow$ 
& CLIP-Score$\uparrow$ 
& AQ$\uparrow$ 
& IQ$\uparrow$ 
& MS$\uparrow$ 
& DD$\uparrow$ \\ 
\midrule
\multicolumn{13}{l}{\textit{External baselines}} \\
ConsisID~\cite{consistid}  
&  &  &  &  &  & 0.566 & 0.593 & 31.47 & 0.526 & 0.670 & 0.982 & 0.51 \\
Phantom~\cite{phantom}     
&  &  &  &  &  & 0.537 & 0.627 & 31.55 & 0.532 & 0.668 & 0.982 & 0.87 \\
VACE~\cite{vace}           
&  &  &  &  &  & 0.423 & 0.444 & 32.61 & 0.564 & \textbf{0.688} & 0.986 & 0.71 \\
IPVG-STD~\cite{jiface}     
&  &  &  &  &  & 0.492 & 0.519 & 31.82 & 0.561 & 0.683 & 0.986 & 0.88 \\
\rowcolor{TableAccent}
\textbf{ST-DRC (Ours)}
&  &  &  &  &  
& \textbf{0.631} & \textbf{0.671} & \textbf{33.04} & \textbf{0.576} & 0.682 & \textbf{0.992} & \textbf{0.93} \\
\midrule
\multicolumn{13}{l}{\textit{Cumulative ablations}} \\
LTX-2.3-Base~\cite{ltx-2}
&  &  &  &  &  & 0.284 & 0.317 & 32.18 & 0.558 & 0.681 & 0.985 & 0.87 \\
+ Ref. Concat. (Sec.~\ref{sec31})
& \checkmark &  &  &  &  & 0.541 & 0.579 & 31.76 & 0.548 & 0.674 & 0.961 & 0.83 \\
+ TASS-RoPE (Sec.~\ref{sec32})
& \checkmark & \checkmark &  &  &  & 0.573 & 0.612 & 31.94 & 0.555 & 0.677 & 0.972 & 0.86 \\
+ Ref. Aug. (Sec.~\ref{sec33})
& \checkmark & \checkmark & \checkmark &  &  & 0.592 & 0.631 & 32.12 & 0.563 & 0.679 & 0.979 & 0.89 \\
+ Aux. ID Loss (Sec.~\ref{sec33})
& \checkmark & \checkmark & \checkmark & \checkmark &  & 0.618 & 0.658 & 32.28 & 0.568 & 0.680 & 0.986 & 0.88 \\
\rowcolor{TableAccent}
+ Decoupled CFG (Sec.~\ref{sec34})
& \checkmark & \checkmark & \checkmark & \checkmark & \checkmark 
& \textbf{0.631} & \textbf{0.671} & \textbf{33.04} & \textbf{0.576} & 0.682 & \textbf{0.992} & \textbf{0.93} \\
\bottomrule
\end{tabular}
\label{tab:exp}
\end{table*}

\section{Experiments}

\subsection{Implementation Details}

\noindent\textbf{Training.}
We build ST-DRC on LTX-2.3~\cite{ltx-2}, and only activate its video branch. 
The model is fully fine-tuned on VIP-200K~\cite{ipvg} for $20$K optimization steps with a batch size of $32$ on H20 GPUs. 
We use AdamW~\cite{adamw} with a learning rate of $5\times10^{-5}$. 
For the auxiliary identity losses in Sec.~\ref{sec33}, we set $\lambda_{\mathrm{id}}=0.1$ and $\lambda_{\mathrm{tic}}=0.05$, and use $w_{\mathrm{aux}}(t)=\left(\frac{\mathrm{SNR}(t)}{\mathrm{SNR}(t)+1}\right)^{\gamma}$ with $\gamma=1.0$. \\
% For decoupled text-reference guidance in Sec.~\ref{sec34}, we use $s_y=5.0$ and $s_r=7.5$ unless otherwise specified.
\noindent\textbf{Metrics.}
We evaluate generated videos from \textbf{\textit{three}} aspects: \textbf{\textit{1)}} identity preservation, \textbf{\textit{2)}} prompt alignment, and \textbf{\textit{3)}} video quality. 
For identity preservation, following ConsisID~\cite{consistid}, we report FaceSim-Arc~\cite{arcface} and FaceSim-Cur~\cite{curricularface}, which compute the facial feature similarity between generated frames and the reference image. 
For text relevance, we use CLIP-Score~\cite{radford2021clip} to measure the alignment between generated videos and input prompts. 
For video quality, following VBench~\cite{vbench}, we report Aesthetic Quality (AQ; LAION aesthetic predictor~\cite{aesthetic-predictor}), Imaging Quality (IQ; MUSIQ~\cite{ke2021musiq}), Motion Smoothness (MS; AMT~\cite{amt}), and Dynamic Degree (DD; RAFT~\cite{raft}).

\subsection{Quantitative and Qualitative Analysis}
\noindent\textbf{Baselines.}
We compare ST-DRC with representative IPVG and S2V methods, including Phantom~\cite{phantom}, VACE~\cite{vace}, ConsisID~\cite{consistid}, and IPVG-STD~\cite{jiface}, covering different reference-conditioning paradigms. 

\noindent\textbf{Quantitative Comparison.}
Tab.~\ref{tab:exp} reports quantitative results on the VIP-200K test set. 
ST-DRC achieves the best identity preservation, improving FaceSim-Arc/Cur from $0.566/0.593$ of ConsisID and $0.537/0.627$ of Phantom to $0.631/0.671$. 
This shows that our latent reference conditioning and face-guided supervision preserve fine-grained facial identity more effectively. 
ST-DRC also obtains the highest CLIP-Score, indicating stronger prompt alignment. 
For video quality, our method achieves the best AQ, MS and DD
, while maintaining competitive IQ. 
Although VACE obtains slightly higher IQ, its identity similarity is much weaker, suggesting that frame-level quality alone is insufficient for facial IPVG. 
Overall, ST-DRC provides the best balance among identity preservation, prompt alignment, temporal smoothness, and visual quality.

\noindent\textbf{Qualitative Comparison.}
Qualitative comparisons are shown in Fig.~\ref{fig:compare}. 
ConsisID suffers from severe copy-paste artifacts and produces noticeable visual artifacts. 
Phantom shows lower visual quality, such as over-exposed or washed-out appearance, and limited temporal identity consistency across frames. 
VACE generates visually plausible videos but exhibits weaker identity similarity to the reference subject. 
IPVG-STD also suffers from noticeable copy-paste behavior, where non-face environmental details from the reference image are often transferred to the generated video. 
In contrast, ST-DRC preserves the target identity more faithfully while maintaining better temporal consistency and video quality.

\subsection{Ablation Study}
We conduct cumulative ablations in Tab.~\ref{tab:exp} to analyze five components of ST-DRC: \textbf{\textit{1)}} reference concatenation, \textbf{\textit{2)}} TASS-RoPE, \textbf{\textit{3)}} reference image augmentation, \textbf{\textit{4)}} auxiliary identity loss, and \textbf{\textit{5)}} decoupled text-reference guidance. 
Starting from the LTX-2.3 video branch, reference concatenation introduces low-level identity cues, but also brings appearance entanglement and copy-paste artifacts. 
TASS-RoPE improves identity retrieval by making reference tokens temporally accessible yet spatially non-overlapping, thereby reducing direct appearance copying. 
Reference augmentation further suppresses pose, color, and background shortcuts, while the auxiliary identity loss provides explicit face-level supervision to improve identity similarity and reduce cross-frame identity drift. 
Finally, decoupled text-reference guidance balances prompt adherence and reference fidelity at inference time. 
Overall, these cumulative gains show that the components are complementary, and the full model achieves the best balance among identity preservation, temporal consistency, text alignment, and video quality.

\section{Conclusion}

We introduced ST-DRC, a spatial-temporal decoupled reference conditioning framework for identity-preserving text-to-video generation. 
ST-DRC represents the reference image as a latent in-context condition and uses TASS-RoPE to keep reference tokens temporally accessible but spatially non-overlapping, enabling identity retrieval while suppressing copy-paste shortcuts. 
With reference-robust identity enhancement and decoupled text-reference guidance, our method improves identity preservation, prompt adherence, and video quality without additional identity encoders or per-identity tuning. 
Overall, ST-DRC achieves strong identity preservation and high overall video quality with a lightweight design, demonstrating its effectiveness for identity-preserving text-to-video generation.

\bibliographystyle{ACM-Reference-Format}
\bibliography{sample-base}

\end{document}